\documentclass[letterpaper, 10 pt, conference]{ieeeconf}  

\IEEEoverridecommandlockouts                              

\usepackage{amsmath}
\usepackage{ctable}
\usepackage{courier}

\usepackage[flushleft]{threeparttable}
\usepackage{multirow}
\newcommand{\figref}[1]{Fig.\,\ref{#1}}
\newcommand{\Figref}[1]{Figure~\ref{#1}}
\newcolumntype{?}{!{\vrule width 1pt}}

\title{\LARGE \bf
Lane-Attention: Predicting Vehicles' Moving Trajectories \\ by Learning Their Attention Over Lanes
}

\author{Jiacheng Pan\textsuperscript{1,2}, Hongyi Sun\textsuperscript{1}, Kecheng Xu\textsuperscript{1}, Yifei Jiang\textsuperscript{1}, Xiangquan Xiao\textsuperscript{1}, Jiangtao Hu\textsuperscript{1}, and Jinghao Miao\textsuperscript{1}
\thanks{\textsuperscript{1} Authors are with Baidu USA LLC, 1195 Bordeaux Drive, Sunnyvale, CA 94089, USA. (\texttt{jiachengpan@gmail.com})}
\thanks{\textsuperscript{2} Now with Google Research.}
}

\begin{document}

\maketitle
\thispagestyle{empty}
\pagestyle{empty}

\begin{abstract}

Accurately forecasting the future movements of surrounding vehicles is essential for safe and efficient operations of autonomous driving cars.
This task is difficult because a vehicle's moving trajectory is greatly determined by its driver's intention, which is often hard to estimate.
By leveraging attention mechanisms along with long short-term memory (LSTM) networks, this work learns the relation between a driver's intention and the vehicle's changing positions relative to road infrastructures, and uses it to guide the prediction.
Different from other state-of-the-art solutions, our work treats the on-road lanes as non-Euclidean structures, unfolds the vehicle's moving history to form a spatio-temporal graph, and uses methods from Graph Neural Networks to solve the problem.
Not only is our approach a pioneering attempt in using non-Euclidean methods to process static environmental features around a predicted object, our model also outperforms other state-of-the-art models in several metrics.
The practicability and interpretability analysis of the model shows great potential for large-scale deployment in various autonomous driving systems in addition to our own.

\end{abstract}

\section{Introduction}
Autonomous driving is a revolutionary technology to free people from tedious and repetitious driving tasks.
During operation, an autonomous driving system repeatedly performs the following four tasks at a high frequency: perceiving the surrounding environment, predicting the possible movements of adjacent objects, planning the ego vehicle's motions, and controlling itself to follow them.
Trajectory prediction of surrounding vehicles plays a crucial role in the overall system because the ego vehicle relies on it to calculate a safe and comfortable moving trajectory.

However, accurately predicting a moving object's trajectory is a challenging task.
Unlike many other sequence prediction problems where a sequence's future states can be inferred merely based on its own historical and current states (e.g. in our case, vehicle past trajectory, and turn signal status, etc.), an object's moving trajectory can be greatly affected by many other external factors, which can be categorized into two types:
1) surrounding static environment, such as
landscapes, lane-lines, and road shapes in the vicinity of the predicted object, and 
2) surrounding dynamic environment, such as
moving objects next to the predicted one and social interaction among them.

\begin{figure}[t]
\centering
\includegraphics[width=0.7\columnwidth]{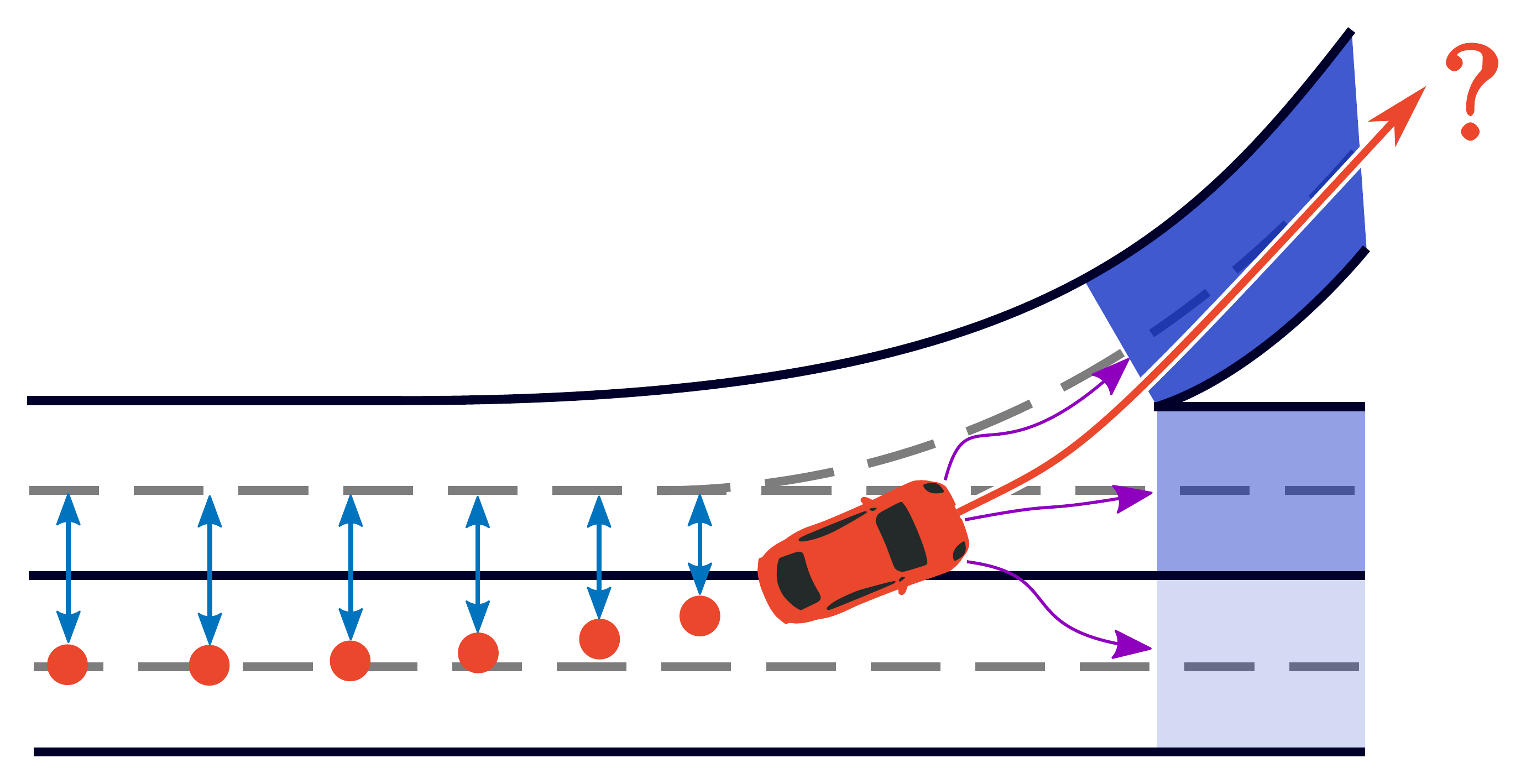}
\caption{Predicting a vehicle's future behaviors by learning its relation to the surrounding lanes.}
\label{fig:static_env}
\end{figure}

There have been extensive researches on learning the dynamic interaction and using it to guide the prediction.
Most of them aim at forecasting pedestrians' trajectories in crowded scenarios.
Early works tackled the problem in a Euclidean (i.e. grid-like or sequence-like) way, by dividing the space into grids and applying occupancy grid pooling or social pooling \cite{social_lstm}; these works were soon superseded by non-Euclidean methods that treated the objects and their interaction as a graph and used attention mechanisms \cite{social_attention} or other methods found in Graph Neural Networks (GNN) to exploit the pairwise interaction.

However, moving vehicles' behaviors, especially on the long run, are much more constrained by lane information (\figref{fig:static_env}), rather than by vehicle dynamics or the occasional interaction with adjacent cars.
Therefore, the impact of static environment can be dominant in determining a vehicle's future moving trajectory, as also indicated by the results and analyses of \cite{argo_paper}.
There have been fewer works in the studies of static environment's influence on vehicle trajectory prediction.
Also, existing state-of-the-art solutions treated road infrastructures as Euclidean data (e.g. semantic map \cite{uber_semantic_map}), which might not necessarily capture their essence due to the following observations:

\begin{itemize}
\item
The structure of lanes on roads is not uniform.
There can be \emph{any} number of lanes around a predicted vehicle, ranging from one to some great number (e.g. when entering a big intersection with many branches).
Also, the shapes or directions of lanes are very \emph{diverse}: on high-ways, lanes are mostly straight; whereas within intersections, lanes may branch into several completely different directions.
\item
While driving, people have their attentions on one or a few of the lanes based on their intention.
They tend to follow, if not exactly, the direction of those lanes.
\end{itemize}

\begin{figure}[t]
\centering
\includegraphics[width=0.99\columnwidth]{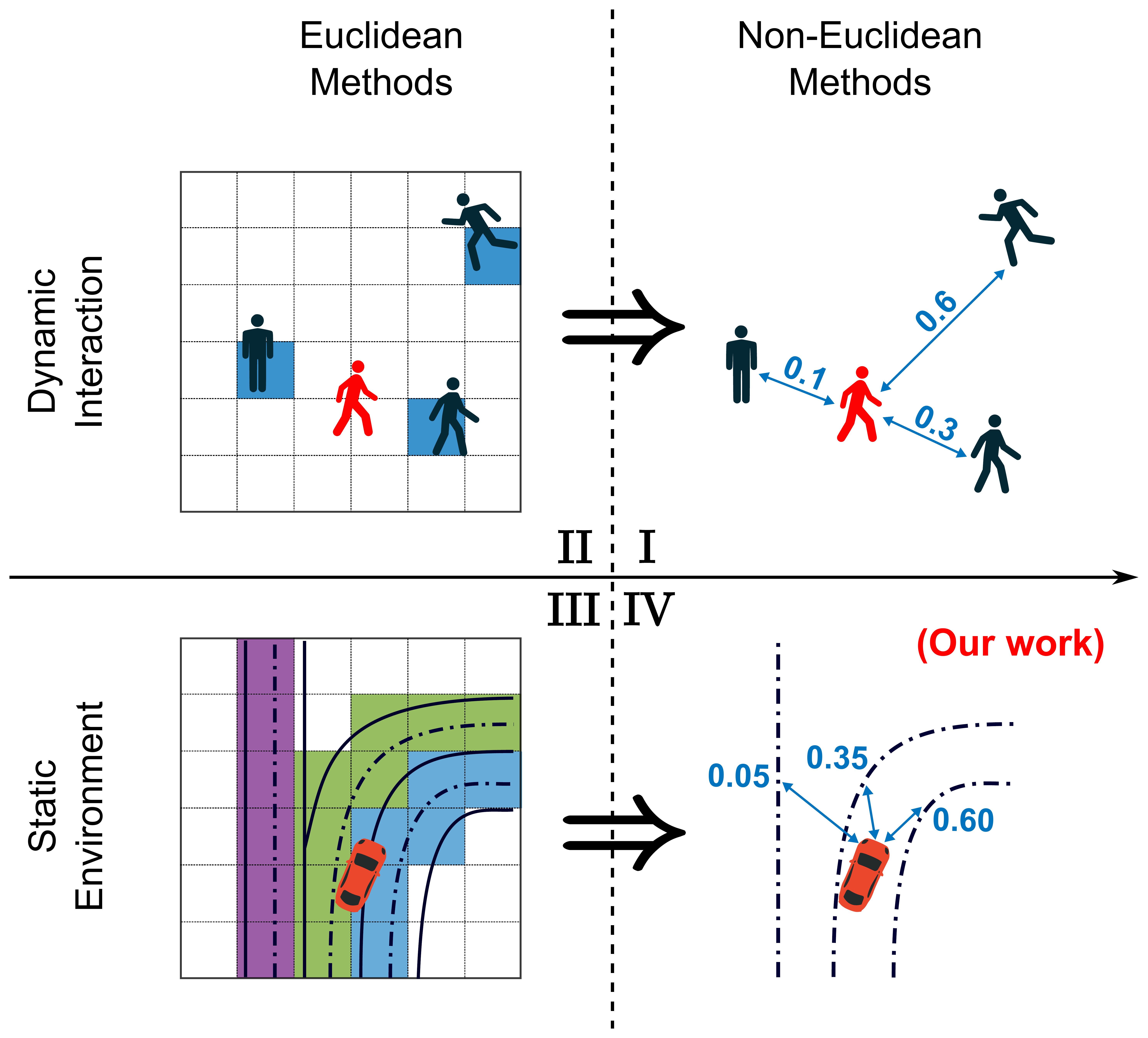}
\caption{The evolution from grid-like processing to non-Euclidean methods (II $\rightarrow$ I) has enabled better modeling of the dynamic interaction. We aim at improving the modeling of static environment in the same way (III $\rightarrow$ IV).}
\label{fig:quadrant}
\end{figure}

We focus on improving the accuracy of vehicle's trajectory prediction through better modeling of static environment's influence.
Inspired by the aforementioned approaches \cite{social_attention} to analyze dynamic environment in terms of pairwise interaction, and motivated by the above observations that pairwise relations among a vehicle and its surrounding lanes play significant roles in predicting the vehicle's future movements, we propose a novel method, the Lane-Attention Neural Network, that treats the lanes as a graph and uses attention mechanisms to aggregate the static environmental information, so that we can achieve successful forecasting of vehicles' moving trajectories (\figref{fig:quadrant}).

Our model (Quadrant IV of \figref{fig:quadrant}) has the following novelties and advantages:
\begin{itemize}
\item It is a \emph{pioneering} attempt to model the static environments using Graph Neural Networks, and its ability to better learn the relation among vehicles and lanes has been proven by the more accurate prediction results than other state-of-the-art works.
We hope this embarkment on a new area can enlighten more upcoming attempts to further improve the understanding and modeling of the influence from surrounding environment on a predicted object.
\item Our solution is \emph{adoptable} to different autonomous driving solutions without additional cost:
our approach can be applied to both high definition (HD) map and non-HD map based autonomous driving.
Note that in the HD map based autonomous driving, the lane information is provided by the stored HD maps; in the non-HD map based autonomous driving, for example, we could leverage camera-detected lanes combined with a few pre-collected human driving paths as lane-structure information.
\item As will be shown, by visualizing the learned attention scores, it can be seen that our algorithm, rather than being a black box itself, provides intuitive explanations of its behaviors.
This great \emph{interpretability} can also benefit other downstream modules of an autonomous driving system. 
\end{itemize}

\section{Related Work}
\subsection{Traditional Models}
Many works used traditional models to predict vehicles' moving trajectories.
Some models, e.g. kinematic models \cite{const_vel1} and dynamic models \cite{vehicle_dynamics}, based the prediction purely on the observed motion history.
Kalman Filter \cite{kalman_filter} has been widely adopted to account for uncertainties in prediction.
Some works used Logistic Regression \cite{intention_logreg}, Support Vector Machine \cite{intention_svm}, or Hidden Markov Model \cite{intention_hmm} to consider a driver's maneuver intention. There have also been attempts \cite{vehicle_interaction} to model interaction among vehicles.

\subsection{Sequence Prediction (Euclidean Methods)}
Great progress has been made in deep neural networks (DNN) in the recent years.
Recurrent Neural Networks (RNN), as well as their variants Long Short-Term Memory (LSTM) \cite{LSTM}
and Gated Recurrent Units (GRU) \cite{GRU}, are good at learning the temporal relations among input features.
They have achieved excellent performance in sequence prediction tasks, such as speech recognition \cite{speech_recog}, machine translation \cite{bahdanau2014neural}, and trajectory prediction \cite{LSTM_traj_pred}, etc.
There have also been attempts \cite{scene_cnn_lstm} to combine Convolutional Neural Networks (CNN) and LSTM for trajectory prediction.

\begin{figure*}[t]
\centering
\includegraphics[width=1.7\columnwidth]{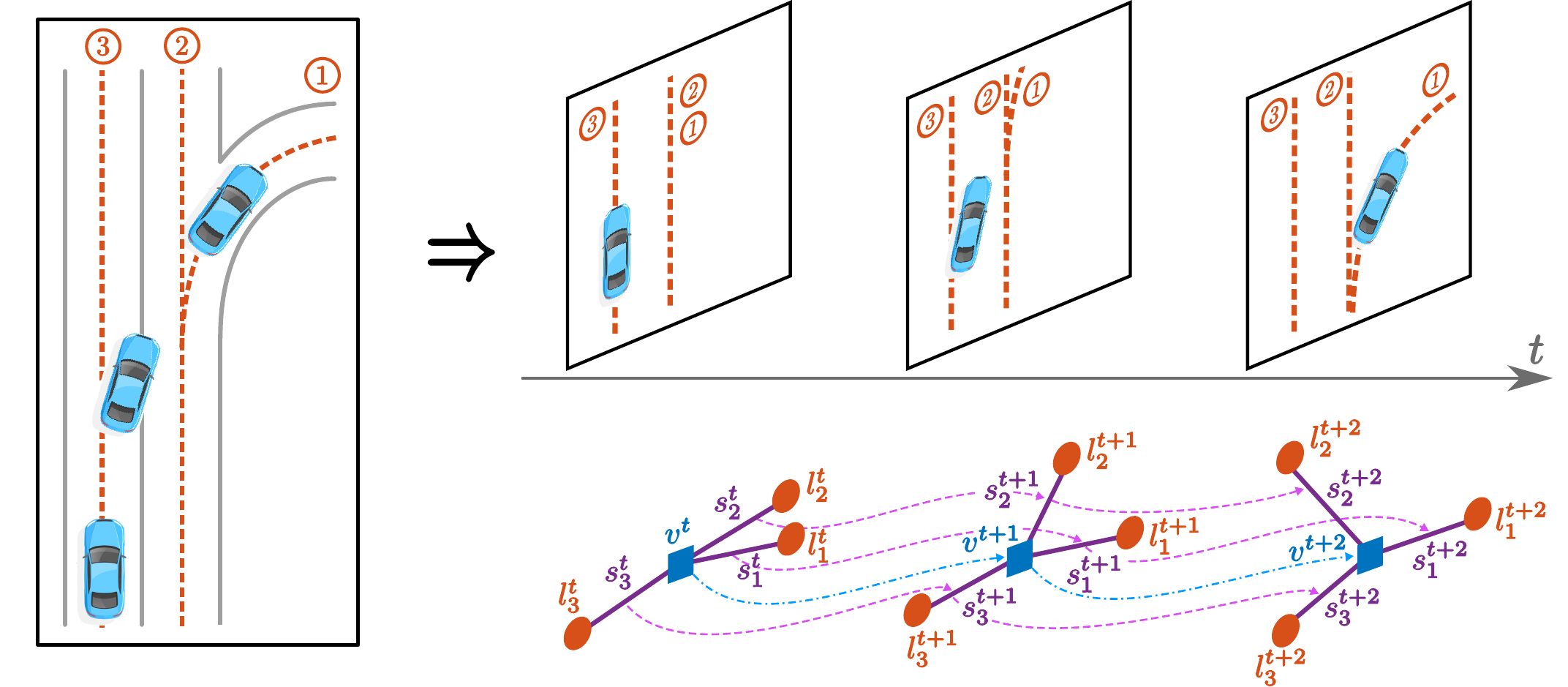}
\caption{Unfolding the history of vehicle's motion on lanes formulates a spatio-temporal graph. Note that in the given example, before their bifurcation, lane 1 and lane 2 are treated as two separate nodes with the same features.}
\label{fig:graph_schematics}
\end{figure*}

\subsection{Graph Neural Networks (GNN)}
RNNs and CNNs work well in extracting features from Euclidean data (those with natural orderings like images or texts), because they impose strong relational inductive biases of locality and translational invariance in time and space \cite{graph_theory}.
In parallel, there exists another class of networks, Graph Neural Networks (GNN) \cite{GNN_early}, that are more effective in handling non-Euclidean inputs or excavating the pairwise relation properties out of input data.
GNN and its variants, such as Graph Convolution Networks \cite{MPNN}, are proven useful not only in processing nonuniform data such as social networks \cite{Graph_SAGE} or knowledge graphs \cite{knowledge_graph}, but also in tasks like object detection \cite{obj_detect_GNN} and neural machine translation \cite{attention_NLP}.

The Spatio-Temporal Graph Neural Networks (ST-GNN) \cite{structural_rnn}, a derivative of GNN, use nodes to represent entities and two kinds of edges to represent temporal and spatial relations.
ST-GNNs find applications in robotics \cite{GNN_physics_engine}, and in many other tasks that require both spatial and temporal reasonings \cite{GNN_object_relation}.
Our work gains inspiration from ST-GNNs.

\subsection{Modeling Social Interactions}
In \cite{social_force}, interaction among pedestrians was modeled by hand-crafted social forces.
Later, \cite{lstm_grid_map} used fine occupancy-grid maps to represent neighboring objects and applied LSTM to learn social interaction among them.
Differently, Social-Pooling \cite{social_lstm} \cite{car_social_pooling} applied coarse grids and used pooling layers to aggregate the neighbor information.
These methods belong to the Quadrant II of \figref{fig:quadrant}.

On the other hand, Social-GAN \cite{social_gan}, corresponding to the Quadrant I of \figref{fig:quadrant}, used Max-Pooling as the symmetric function \footnote{A symmetric function takes any number of inputs but generates a fixed-dimension output.} to aggregate all neighbor information.
Similarly, Social-Attention \cite{social_attention} and SR-LSTM \cite{sr_lstm} formulated the problem as ST-Graphs and utilized attention mechanisms.
TrafficPredict \cite{TrafficPredict} used a ST-Graph with multiple node categories to model various relations among different types of traffic participants.

\begin{figure*}[t]
\centering
\includegraphics[width=2.0\columnwidth]{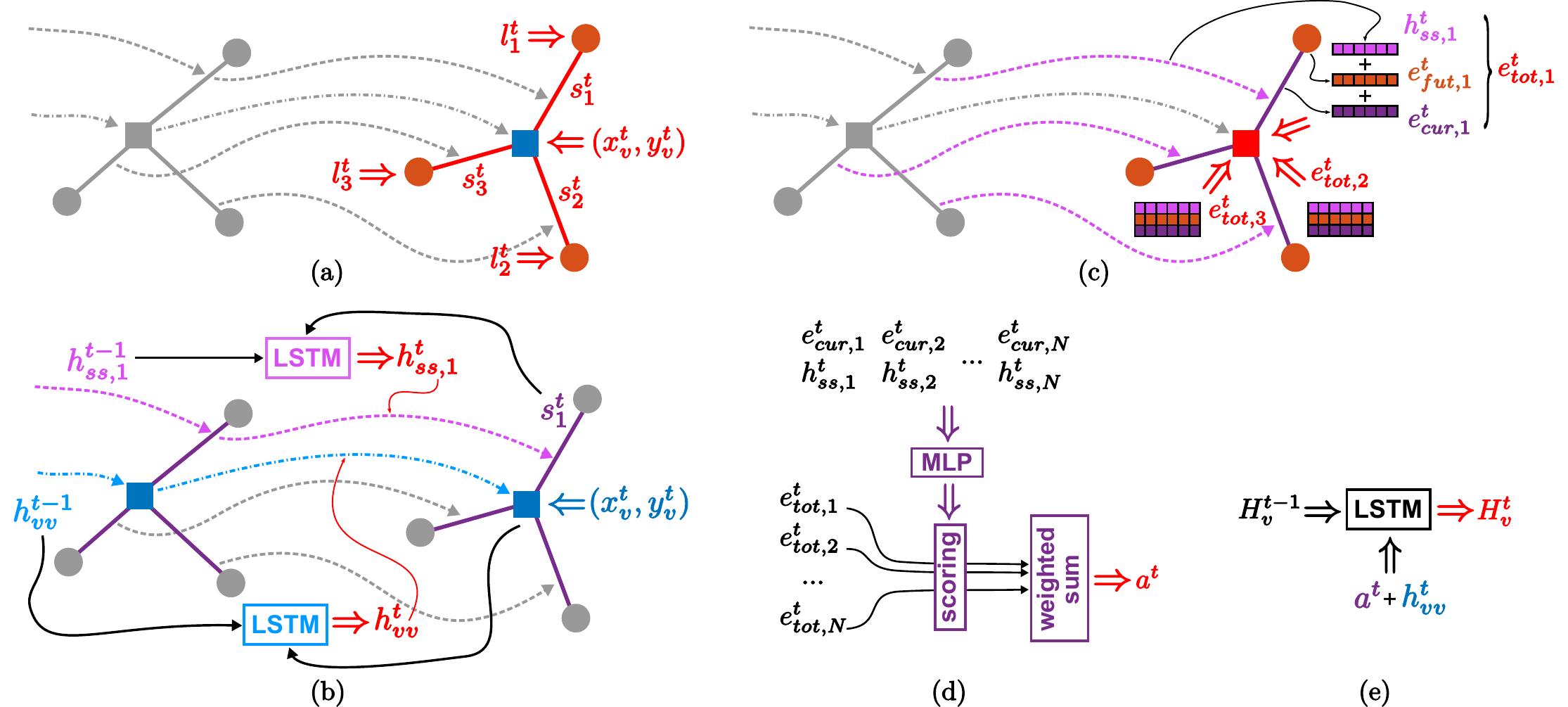}
\caption{At every time-step, there will be (a) reception of new information, (b) temporal evolution to update $\mathcal{E}_T$, (c) spatial aggregation to update $v_t$, details of which (Lane-Attention) are shown in (d), and (e) updates of the overall state.}
\label{fig:graph_details}
\end{figure*}

\subsection{Modeling Static Environments}
To model the static environment, Scene-LSTM \cite{scene_lstm}, SS-LSTM \cite{ss_lstm}, and other works \cite{scene_cnn_lstm} \cite{lane_cnn} applied CNN to a bird's-eye view photo of the environment, directly or after some preprocessing.
Alternatively, the inputs to CNN could be semantic maps \cite{uber_semantic_map}, processed from pre-collected HD-maps, with a variety of colors representing different lane directions and with fading rectangles to capture vehicle movement history.
One recent work \cite{argo_paper} projected a predicted vehicle onto a given lane, and used the lateral and longitudinal displacements as input features.
These methods all fall in the Quadrant III of \figref{fig:quadrant}, and our work explores their missing counterpart in the Quadrant IV.

\section{Methods}

\subsection{Problem Definition}
We receive as inputs each vehicle's historical positions from $t=-T_{obs}$ to the current time-step $t=0$, at increments of $\Delta t$, the sampling period of sensors.
It is also assumed that at each time-step, every vehicle's surrounding lanes are given, and the number of lanes is denoted as $N$.
Our goal is to predict each vehicle's future positions over a time-span of $T_{pred}$, which should be an integer multiple of $\Delta t$.
To avoid cumbersome indexing, all the notations below refer to an arbitrary single vehicle instance out of all the input data.

\subsection{Spatio-Temporal Graph (ST-Graph) Formulation}
To clearly manifest pairwise relations, we formulate the problem as a spatio-temporal graph: $\mathcal{G}=(\mathcal{V}, \mathcal{E}_T, \mathcal{E}_S)$, where $\mathcal{V}$ is the set of nodes, $\mathcal{E}_T$ is the set of temporal edges, and $\mathcal{E}_S$ is the set of spatial edges.
\begin{IEEEeqnarray}{rCl}
\mathcal{V} &=& \{v^{t}, l_{i}^{t}\}\mbox{, and } \mathcal{E}_S = \{s_{i}^{t}\} \mbox{, where } s_{i}^{t} = (v^{t}, l_{i}^{t}) \mbox{,}  \IEEEnonumber\\
&\forall t& \in [-T_{obs}/\Delta t, T_{pred}/\Delta t] \mbox{ and } \forall i \in [1, N] \mbox{.} \label{eq:node_range}\\
\mathcal{E}_T &=& \{(s_{i}^{t}, s_{i}^{t+1}), (v^t, v^{t+1})\} \mbox{,}  \IEEEnonumber\\
&\forall t& \in [-T_{obs}/\Delta t, T_{pred}/\Delta t - 1] \mbox{ and } \forall i \in [1, N] \mbox{.} \label{eq:edge_range}
\end{IEEEeqnarray}
\eqref{eq:node_range} means that $\mathcal{V}$ contains two kinds of nodes: a vehicular node $v^{t}$ represents a vehicle at a given time $t$, and a lane node $l_i^t$ represents one of the local lanes around the predicted vehicle at time $t$.
The pair-wise relations between $v^t$ and $l_i^t$ at the same $t$ form the set of spatial edges $\mathcal{E}_S$.
\eqref{eq:edge_range} indicates that there are two types of temporal edges, one about the vehicle's state evolution and the other about the evolution of lane-vehicle relationship over time.
In a nutshell, it could be seen as the vehicle's movement history, as well as its changing relation with the surrounding lanes, is unrolled over time to form an ST-graph (\figref{fig:graph_schematics}).

At every instant, $v^t$ receives the vehicle's new spatial position $(x_v^t, y_v^t)$; $l_i^t$ is also refreshed to reflect lanes in the vehicle's current neighborhood.
Typically $l_i^t$ contains a set of ordered lane-points.
The lane information can come directly from the sensed and perceived lane-lines.
Alternatively, it can be derived by first localizing the vehicle's position, and then fetching the lanes around it from a pre-collected HD map.
With these new features, all the $s_i^t \in \mathcal{E}_S$ of this instant are then readily updated with the vehicle's new spatial relation to its local lanes (\figref{fig:graph_details}\,(a)).

Next, there will be temporal evolution to update $\mathcal{E}_T$ and spatial aggregation to update $v_t$ of this time-step, with details covered in the following sub-sections.

\subsection{Temporal Evolution}
\subsubsection{Vehicle State Evolution}
Long Short-Term Memory networks (LSTM) have been successful in learning the patterns of sequential data.
A standard LSTM network can be described by the following equations:
\begin{align}
f_t &= \sigma(W_f x_t + U_f h_{t-1} + b_f) \mbox{,}\label{eq:LSTM_eq_1}\\
i_t &= \sigma(W_i x_t + U_i h_{t-1} + b_i) \mbox{,}\\
o_t &= \sigma(W_o x_t + U_o h_{t-1} + b_o) \mbox{,}\\
c_t &= f_t \odot c_{t-1} + i_t \odot \tanh(W_c x_t + U_c h_{t-1} + b_c) \mbox{,}\\
h_t &= o_t \odot \tanh(c_t) \mbox{,}\label{eq:LSTM_eq_5}
\end{align}
where $f_t$, $i_t$, $o_t$, and $c_t$ stand for forget gate, update gate, output gate, and cell state, respectively.
$h_t$ is the hidden state and contains encoded patterns of the sequential inputs.
We will use
\begin{align}
h_t = \mbox{LSTM}(h_{t-1}, x_t; \Theta)
\end{align}
for the rest of the paper as the abbreviation of \eqref{eq:LSTM_eq_1} -- \eqref{eq:LSTM_eq_5}.

A vehicle's movement is a form of sequential data, and it is in part governed by, especially in short term, kinematics and vehicle dynamics.
For example, a vehicle can't complete a sharp turn instantaneously; nor can it slow down from 60\,mph to 0 in a blink.
Therefore, we use a standard LSTM network to learn this underlying driving force:
\begin{align}
e_{vv}^t &= \mbox{MLP}((x_v^t - x_v^{t-1}, y_v^t - y_v^{t-1}); W_{vv}) \mbox{,} \label{eq:e_vv}\\
h_{vv}^t &= \mbox{LSTM}(h_{vv}^{t-1}, e_{vv}^t, \Theta_{vv}) \mbox{.} \label{eq:h_vv}
\end{align}
The network first embeds the relative displacement using a Multi-Layer Perceptron (MLP) network as in \eqref{eq:e_vv}, and then uses the embedding and the previous hidden state as inputs to update the new hidden state for the temporal vehicle-to-vehicle edge $(v_t, v_{t+1})$ as in \eqref{eq:h_vv} (\figref{fig:graph_details}\,(b)).

\subsubsection{Lane-Vehicle Relation Evolution}
In addition to the laws of physics, what's also determining a vehicle's movement is the driver's intention.
One's intention is often not expressed explicitly, but can be inferred based on the vehicle's changing relation with each lane because drivers tend to follow one or a few lanes to stay courteous and to avoid accidents.
We capture this relation with another LSTM network.

First, with the vehicle's new position $(x_v^t, y_v^t)$ and the updated local lane information $l_i^t$, we project the vehicle's location onto each lane to get a projection point $(x_{p,i}^t, y_{p,i}^t)$.
Then, we get the difference between projection points and vehicle position, and use MLP to embed this vector: \eqref{eq:e_ss}.
Finally, as shown in \eqref{eq:h_ss}, this embedding and the previous hidden state are used to update the new hidden state $h_{ss,i}^t$, which corresponds to the temporal edge $(s_i^t, s_i^{t+1})$ connecting sequential lane-vehicle relation pairs (\figref{fig:graph_details}\,(b)).
\begin{align}
e_{ss,i}^t &= \mbox{MLP}((x_{p,i}^t - x_v^t, y_{p,i}^t - y_v^t); W_{ss}) \mbox{,} \label{eq:e_ss}\\
h_{ss,i}^t &= \mbox{LSTM}(h_{ss,i}^{t-1}, e_{ss,i}^t, \Theta_{ss}) \mbox{.} \label{eq:h_ss}
\end{align}
$h_{ss,i}^t$ is expected to contain the learned evolving relation between a vehicle and the i$^{th}$ lane.
We will next show how this hidden state, as well as other information, of all lanes can be aggregated to infer a driver's intention and accurately predict the vehicle's future trajectory. 

\subsection{Spatial Aggregation}
For each lane, we have an encoding $h_{ss,i}^t$ of its historical evolving relation with the vehicle.
We can further encode its current relative position to the vehicle and its future shape, each using an MLP network:
\begin{align}
e_{cur,i}^t &= \mbox{MLP}((x_{p,i}^t - x_v^t, y_{p,i}^t - y_v^t); W_{cur})  \mbox{,} \\
e_{fut,i}^t &= \mbox{MLP}(l_i^t; W_{fut}) \mbox{,}
\end{align}
and concatenate all three vectors together to form $e_{tot,i}^t$, the overall encoding for each lane at $t$:
\begin{align}
e_{tot,i}^t &= \mbox{concatenate}(h_{ss,i}^t, e_{cur,i}^t, e_{fut,i}^t) \mbox{.} \label{eq:e_tot}
\end{align}
To jointly reason across multiple lanes, we must effectively aggregate the encodings of all lanes (\figref{fig:graph_details}\,(c)).
This is a challenging task, because there can be variable number of lanes but the aggregated output should be compact and of fixed dimension.
Also, different lanes play different roles in determining a vehicle's future movement, and the aggregation module needs to take that into consideration too.
Therefore, we tried two different methods for this.

\subsubsection{Lane-Pooling}
The Lane-Pooling method assumes the deciding factor is a single lane.
This single lane is the one that's closest to the vehicle and it may vary over time.
At each time-step, Lane-pooling selects the encoding of the lane that's closest to the vehicle, and uses it as the aggregated encoding $a^t$:
\begin{align}
i_{pooling} &= \underset{i}{\arg\min}((x_{p,i}^t - x_v^t)^2 + (y_{p,i}^t - y_v^t)^2) \mbox{,} \\
a^t &= e_{tot, i_{pooling}}^t \mbox{.} \label{eq:lane_pooling_at}
\end{align}

\subsubsection{Lane-Attention}
However, it may not be the case that a driver only focuses on single lane while driving; the driver may rather pay attention to multiple lanes.
Also, in some cases, such as in the middle of a lane-changing behavior, there will be an abrupt change in the lane-pooling result, and this may introduce some negative impacts on the subsequent network modules.
To resolve the above problems, we developed Lane-Attention.

For the operation of Lane-Attention, first, we compute an attention score for each lane based on its current location and historical relation to the vehicle,
\begin{align}
\mbox{score}(i,t) = \mbox{MLP}((\mbox{concatenate}(e_{cur,i}^t, h_{ss,i}^t)); W_{score}) \mbox{.}
\end{align}
Then, the overall encoding $a^t$ is computed by taking a weighted sum (\figref{fig:graph_details}\,(d)) of each lane's total encoding $e_{tot,i}^t$ from \eqref{eq:e_tot}, with the weights being the normalized attention scores,
\begin{align}
a^t = \sum_{i=1}^{N}\frac{\exp{(\mbox{score}(i,t))}}{\sum_{j=1}^{N} \exp{(\mbox{score}(j,t))}}\cdot e_{tot,i}^t \mbox{.} \label{eq:lane_attn}
\end{align}

The resulting aggregated lane encoding $a^t$, either from Lane-Pooling or from Lane-Attention, is expected to contain learned encoding of a driver's intention.
Next, $a^t$, together with the previous encoding of vehicle's movement history, will be combined and used to update the overall hidden-state corresponding to the vehicular node $v^t$:
\begin{align}
e_{v}^t &= \mbox{concatenate}(a^t, h_{vv}^t) \mbox{,} \\
H_{v}^t &= \mbox{LSTM}(H_{v}^{t-1}, e_{v}^t, \Theta_{v}) \mbox{.}
\end{align}
$H_v^t$ gets updated at every time-step (\figref{fig:graph_details}\,(e)), and can be used to infer a vehicle's future moving trajectory. 

\begin{table*}[!ht]

 \renewcommand{\arraystretch}{1.3}
 \caption{Performance Comparison} \label{performance table}
 \centering
 \begin{threeparttable}
 \begin{tabular}{?c?c?c?c?c?c?c?}
 \specialrule{1pt}{0em}{0em}
 $T_{pred}$ & Metrics & ~~~~~ LSTM ~~~~~ & \,~Semantic Map~\, & ~~ Single-Lane ~~ &  ~~Lane-Pooling~~ &  ~Lane-Attention~ \\
   \specialrule{.2em}{0em}{0em}
\multirow{2}{*}{\shortstack{1 sec.}} & ADE$^{\dagger}$ & 0.2595 & 0.2826 & 0.2286 & 0.2280 & \textbf{0.2238} \\ \cline{2-7}
                                    & FDE$^{\ddagger}$ & 0.4823 & 0.5674 & 0.4097 & 0.4085 & \textbf{0.3979} \\ \specialrule{1pt}{0em}{0em}    
\multirow{2}{*}{\shortstack{3 sec.}} & ADE$^{\dagger}$ & 1.3257 & 1.3970 & 0.9557 & 0.9374 & \textbf{0.9045} \\ \cline{2-7}
                                    & FDE$^{\ddagger}$ & 3.3415 & 3.1792 & 2.2885 & 2.2336 & \textbf{2.1299} \\ \specialrule{1pt}{0em}{0em}                           
 \end{tabular}
\begin{tablenotes}
\footnotesize
\item[$\dagger$] ADE: average displacement error (in meters).
\item[$\ddagger$] FDE: final displacement error (in meters).
\end{tablenotes}
 \end{threeparttable}
 \end{table*}

\subsection{Trajectory Inference and Loss Function}
When predicting the trajectory of each vehicle at time $t \in [1, T_{pred}/\Delta t]$, we assume that each trajectory point follows a bi-variate Gaussian distribution, and we train the network to learn all the parameters of the distribution.
Therefore, we process the hidden states $H_v^t$ of vehicular node using an MLP with the last rectified linear units (ReLU) layer removed, and output a 5-dimensional vector for each trajectory point, containing values of the mean vector and covariance matrix:
\begin{align}
[\mu_x^t, \mu_y^t, \sigma_x^t, \sigma_y^t, \rho^t] = \mbox{MLP}(H_v^t; W_{pred}) \mbox{.}
\end{align}
We then use the expectation of the predicted distribution, ($\mu_x^t, \mu_y^t$), as the delta displacement and add it to $(x_v^{t}, y_v^{t})$ to generate the new spatial position of the vehicle: $(x_v^{t+1}, y_v^{t+1})$, which will serve as the input to the LSTM of the subsequent cycle.
This process is repeated until we finish predicting all the trajectory points up to $t=T_{pred}/\Delta t$.

We use the negative log-likelihood as the loss function and train the network by minimizing this loss:
\begin{align}
L = - \sum_{t=1}^{T_{pred}/\Delta t} \log{(P(x_v^t, y_v^t | \mu_x^t, \mu_y^t, \sigma_x^t, \sigma_y^t, \rho^t))} \mbox{.}
\end{align}

\section{Evaluation}
Our model has been implemented and tested using the Apollo open-source platform \cite{apollo_open}. This section presents the experimental setup and quantitative and qualitative analysis of results.

\subsection{Dataset Description}
We collected traffic data in urban areas using our autonomous vehicles built on Lincoln MKZs, equipped with Velodyne HDL-64E LiDAR and Leopard LI-USB30-AZ023WDRB cameras.
The collected data includes 1) point clouds from LiDARs for object detection and localization; 2) images from cameras for object and lane-line detection. %
The raw data was immediately processed by computer vision algorithms to detect and track objects.
The sampling period $\Delta t$ is 0.1 second for our system.

\begin{figure}[!b]
\centering
\includegraphics[width=0.9\columnwidth]{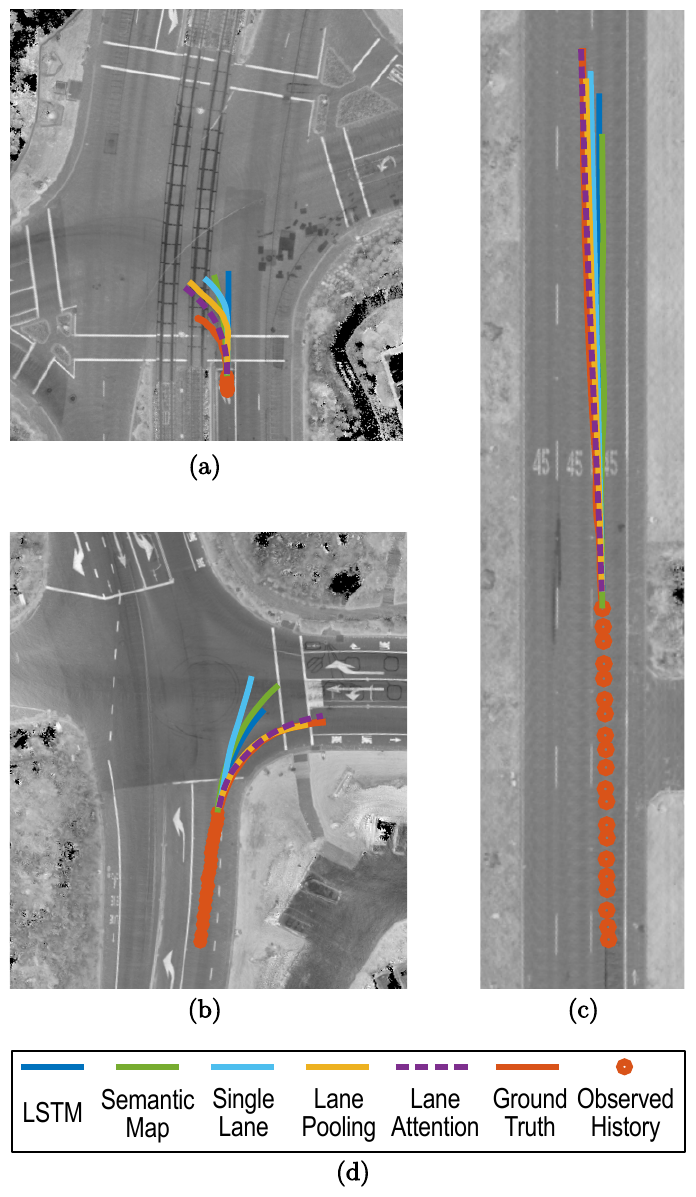}
\caption{A few representative cases showing all models' prediction for (a) left-turning, (b) right-turning, and (c) high-speed driving and lane-changing. Lane-Attention made the best prediction. Legends are shown in (d).}
\label{fig:multi_traj}
\end{figure}

For the detected objects, we filtered out non-vehicular objects and those with less than 3 seconds of tracking.
For each remaining object, we used 3 seconds of trajectory as the ground-truth label and the history right before that (which can contain as small as 0.1 second and up to 2 seconds of records) as input features, for model training and testing.

The resulting dataset contains 870,107 samples.
Among them, 6.2\% are left-turn or U-turn behaviors, 5.9\% are right-turn behaviors, 6.4\% are lane-changing, and the rest 81.5\% are mostly driving along the road, straight or curvy. 
We split them into three sets for training, validation, and testing. 
The partition is based on dates so that no samples from the same day are spread into different sets.
This can greatly decorrelate the data and ensure good generalizability learned by models.
The numbers of samples in training, validation, and testing sets follow a ratio of 6 : 2 : 2.5.

\begin{figure*}[!t]
\centering
\includegraphics[width=2.05\columnwidth]{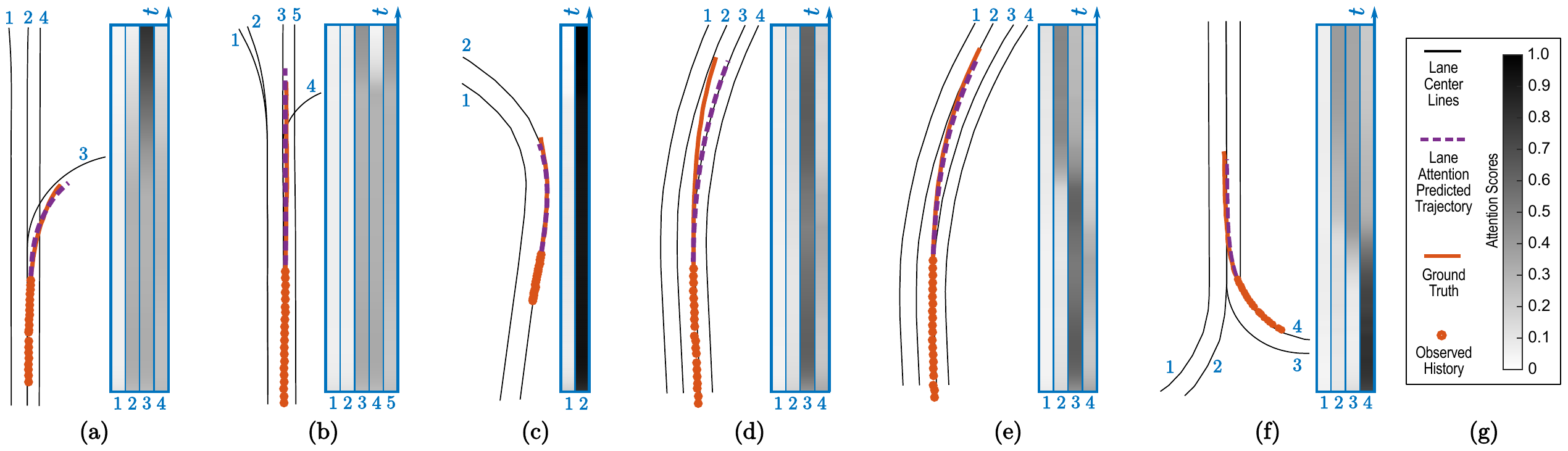}
\caption{(a)-(f) show a few examples of the Lane-Attention model's prediction vs. actual trajectories, and corresponding visualizations of the learned attention for each lane as a function of time. (g) is the legend for them.}
\label{fig:visual_attn}
\end{figure*}

\subsection{Implementation Details}
For the two LSTM networks of \figref{fig:graph_details} (b), the dimensions of embeddings and hidden states are 32 and 64.
All the $h_{ss,i}^t$, $e_{fut,i}^t$, and $e_{cur,i}^t$ of \figref{fig:graph_details} (c) are 64-dimensional vectors.
Therefore, after aggregation in \figref{fig:graph_details} (d), the resulting $a^t$ has a size of 192.
Finally, the combined size of $a^t$ and $h_{vv}^t$ is 256, which is processed by the LSTM of \figref{fig:graph_details} (e) that also uses 256 as the size of hidden states.
The model was trained using Adam with a initial learning rate of 0.0003.
When the validation loss plateaued for more than three epochs, the learning rate was reduced to 0.3$\times$ the previous value.
The entire pipeline was implemented using PyTorch framework and the training was done on a single Nvidia Titan-V GPU.

\subsection{Experimental Results}
We separately trained models to predict 1 second and 3 seconds of future trajectory, and evaluated their performance using the following metrics:
\begin{itemize}
\item Average Displacement Error (ADE): the Euclidean distance between predicted points and ground truth, averaged over the entire predicted time steps.
\item Final Displacement Error (FDE): the Euclidean distance between the predicted position at the final timestamp ($t=T_{pred}$) and the actual final location. This is also the maximum displacement error in most cases, because the models predict delta displacements and errors will accumulate and reach the peak at the end.
\end{itemize}

Besides the Lane-Pooling and Lane-Attention models, we trained three more models for benchmark purposes:
\begin{itemize}
\item LSTM: A simple LSTM that considers motion history only, without modeling the surrounding lanes.
\item Semantic Map \cite{uber_semantic_map}: This approach 
used a rasterized semantic map to represent environmental features.
We reproduced the semantic maps, which contain lanes highlighted in different colors indicating their relations (adjacent, connected, or of reverse direction, etc.), intersection and road boundaries, and bounding boxes with fading colors to represent the predicted object's motion history. \footnote{Since code and the original dataset are not available, we implemented the algorithm and trained the model on our own dataset.}
CNNs are used to process the semantic map to help with the trajectory prediction.
\item Single-Lane \cite{argo_paper}: 
This method focuses on a single lane of interest.
We implemented it by selecting the lane based on its proximity to the vehicle at the beginning of the prediction period.
This lane's encoding was treated as the pooled result of \eqref{eq:lane_pooling_at} and the remaining processing was the same as that of the Lane-Pooling method.
\end{itemize}

As indicated by the test results, the Lane-Attention model achieved the best prediction accuracy across all metrics (Table 1).
Also, we note that although the gaps among model performance are relatively small when predicting 1 second of trajectory (e.g. the ADE of LSTM is only 16\% higher than that of Lane-Attention), they get much larger when 3 seconds of future trajecotry are predicted (e.g. the ADE of LSTM is now $\sim$ 1.5$\times$ that of Lane-Attention. So is the FDE.).
This validated our prior expectation that long-term prediction is more heavily dependent on a driver's intention which is better learned by our Lane-Attention Neural Network.
The final displacement error is 0.3979\,m after 1 second. It is 2.1299\,m after 3 seconds, roughly the width or half the length of a mid-size car.

We would also like to point out that, compared with other works \cite{uber_semantic_map}, HD map is not a requirement for our model.
Our model works even with the minimum perception of predicted objects and lane center-lines, without the need to know details like intersection or road boundary, and reverse lane information, etc.
This makes our model feasible for many low-cost pure-visual autonomous driving solutions as well, such as Apollo Lite \cite{apollo}.
The resulting prediction accuracy is also dependent on the sensors used and hardware configurations.

\Figref{fig:multi_traj} shows a few representative cases comparing the prediction from various models.
Among all, Lane-Attention achieved the closest forecasting to the actual trajectories.
The model also demonstrates strong de-noising capability.
For example, as in \Figref{fig:multi_traj}\,(c), even with measurement errors in longitudinal displacements, the model still achieved an accurate prediction for a lane-changing vehicle.

\subsection{What Has the Model Learned?}
It is of great interest to see what has the model learned to accomplish the great performance.
One tangible way is to visualize the learned attention scores on various lanes as functions of time, and a few exemplary cases are shown in \figref{fig:visual_attn}.
We make a few observations:
\begin{itemize}
\item As indicated by \figref{fig:visual_attn} (a) and (b), the model has learned to gradually shift its attention away from lanes that are becoming irrelevant and focus on the really significant ones which the driver intends to follow.
\item From the comparison between (a)(b) and (c) of \figref{fig:visual_attn}, it could be seen that the model learned to focus on multiple lanes ahead while driving straight, but pay high amount of attention to the edge lane if following curvy roads, quite similar to what a human driver would do.
\item There are cases when our prediction deviates from the ground truth (\figref{fig:visual_attn} (d)).
A significant number of such cases happen when a maneuver is done at some future time and there is no sign of that at the moment.
Even human drivers cannot make correct predictions for these scenarios.
However, whenever such sign appears, even if it is inconspicuous, our model will correctly predict the future trajectory as in \figref{fig:visual_attn} (e) (a few hundred milliseconds after \figref{fig:visual_attn} (d)).
Also, \figref{fig:visual_attn} (e) indicates that during lane-changing, our model gradually shifts the attention from the vehicle's original lane to the target one.
\item In addition to lane bifurcation cases as in \figref{fig:visual_attn} (a) and (b), the model has also learned to predict in cases of lane merging (\figref{fig:visual_attn} (f)). Note that lanes have the same attention scores after merging, which makes sense as they become one lane.
\end{itemize}
In summary, our model has learned to infer human drivers' intention.
This learned results (e.g. attention scores), in addition to the predicted trajectories, can also be passed to the subsequent planning module of an autonomous driving system for a more reasonable planning of ego vehicle's behaviors, on which will be elaborated by our future works.

\section{Conclusion}
This paper presents a deep neural network that leverages motion history and surrounding environment to predict a vehicle's moving trajectory.
By formulating the task as a spatio-temporal graph, using LSTM-based temporal evolution, and applying spatial aggregation of attention mechanisms, our model has been trained to learn drivers' intention, manifested as the different levels of attention scores.
The evaluation of our model's performance on road-test data collected in real environment has demonstrated its ability to predict trajectories that are highly representative of real ones, as well as its better prediction accuracy than existing models implementing Euclidean techniques.

Our model, as indicated by Quadrant IV of \figref{fig:quadrant}, tries to fill in the missing piece of modeling static environment using non-Euclidean methods.
Combining it with various methods in Quadrant I of \figref{fig:quadrant} can enable a more robust prediction in complicated environment with congested traffics.
Furthermore, with advancement in V2X \cite{apollo_v2x}, limits like an autonomous vehicle being unable to perceive traffic lights seen by other vehicles can be overcome.
This traffic light feature can be considered to enhance model performance in the future.

\bibliographystyle{IEEEtran}
\bibliography{ref}

\end{document}